\documentclass[runningheads]{llncs}
\usepackage{graphicx}
\usepackage{multirow}
\usepackage{tabularx}
\usepackage[table]{xcolor}
\usepackage{array}
\usepackage{wrapfig}
\usepackage{lipsum} 
\usepackage{caption}
\usepackage{tabulary}
\usepackage[labelfont=bf,justification=raggedright, singlelinecheck=false]{caption}
\usepackage{hyperref}
\definecolor{veryLightGray}{gray}{0.9}
\begin{document}
\title{ Explainers’ Mental Representations of Explainees’ Needs in Everyday Explanations}
\titlerunning{Explainers’ Mental Representations of Explainees’ Needs}
%
%
\author{Michael Erol Schaffer \and
Lutz Terfloth\and Carsten Schulte\and
Heike M. Buhl}
\authorrunning{M. E. Schaffer et al.}
%
\institute{Paderborn University, Paderborn, Germany \\
\email{\{michael.schaffer,lutz.terfloth,carsten.schulte,\\ heike.buhl\}@uni-paderborn.de}\\}
\maketitle              
\vspace{-0.5cm}
\begin{abstract}
In explanations, explainers have mental representations of explainees’ developing knowledge and shifting interests regarding the explanandum. These mental representations are dynamic in nature and develop over time, thereby enabling explainers to react to explainees’ needs by adapting and customizing the explanation. XAI should be able to react to explainees’ needs in a similar manner. Therefore, a component that incorporates aspects of explainers’ mental representations of explainees is required. In this study, we took first steps by investigating explainers’ mental representations in everyday explanations of technological artifacts. According to the dual nature theory, technological artifacts require explanations with two distinct perspectives, namely observable and measurable features addressing “Architecture” or interpretable aspects addressing “Relevance”. We conducted extended semi structured pre-, post- and video recall-interviews with explainers (N=9) in the context of an explanation. The transcribed interviews were analyzed utilizing qualitative content analysis. The explainers’ answers regarding the explainees’ knowledge and interests with regard to the technological artifact emphasized the vagueness of early assumptions of explainers toward strong beliefs in the course of explanations. The assumed knowledge of explainees in the beginning is centered around Architecture and develops toward knowledge with regard to both Architecture and Relevance. In contrast, explainers assumed higher interests in Relevance in the beginning to interests regarding both Architecture and Relevance in the further course of explanations. Further, explainers often finished the explanation despite their perception that explainees still had gaps in knowledge. These findings are transferred into practical implications relevant for user models for adaptive explainable systems.
\end{abstract}
%
\vspace{-0.8cm}
\section{Introduction}
For a long period of time, XAI research focused solely on technological aspects - for example, creating algorithms as a foundation for XAI. In recent XAI
\let\thefootnote\relax\footnotetext{This is a full paper version of: Schaffer, M. E., Terfloth, L., Schulte, C., Buhl, H. M.: Perception and consideration of the explainees’ needs for satisfying explanations. Joint Proceedings of the xAI-2024 Late-breaking Work, Demos and Doctoral Consortium. \textbf{3793}, 17-24 (2024).}


%
\noindent research, this predominantly technological perspective shifted toward a perspective that emphasizes the importance of incorporating human factors in the form of a socio-technological approach \cite{ref_article41}. As research in this regard is still at an early stage \cite{ref_article23}, stepping up efforts to accelerate development of mechanisms that perceive and react to human needs for person-specific and adaptive explainable systems is a necessity.\\
Thus far, numerous studies have not considered if given explanations meet the needs and expectations of end users \cite{ref_article21,ref_article33}. It can be concluded that XAI would benefit from increased consideration of human end users, thereby implying that XAI should recognize these individuals with their specific needs in specific contexts \cite{ref_article21,ref_article33,ref_article34,ref_article38,ref_article41}. These individuals require different explanations \cite{ref_article38}, and XAI should be able to understand why the individual requires a specific explanation \cite{ref_article11}. Enabling XAI to adapt to specific needs and requests will enable customized explanations \cite{ref_article23} that are of greater benefit for the end user. Imagine, for example, a medical setting in which the end user could be a practicing physician, a medical student, or even a patient. Every end user will utilize the XAI system in a slightly different manner and for different reasons, thereby altering the foci and context in which the XAI system is used.\\
According to Ribera and Lapedriza \cite{ref_article38}, user-centered XAI should aim to answer the following questions: Why does something need to be explained? What needs to be explained? How does it need to be explained? And who does it need to be explained to? Imagine someone requires an explanation on how to use an app vs. on how to alter the algorithms of an app. For this, XAI should have a component that contains information regarding human end users as well as their developing knowledge and interests.\\
To be able to develop such a component, it has to be understood how mental representations of humans evolve and which information they contain. Therefore, in this empirical study, we investigated everyday explanations among human interlocutors to understand the role of mental representations the explaining person has of the other interlocutor. It is of special interest to investigate what the explainer learns about the developing knowledge and interests of the other interlocutor. The study aims at identifying aspects within the mental representations that are helpful to develop advanced XAI with the ability to react to the needs of users. This paper also includes practical implications for structuring synthetic explanations. \\
In the following account, we first provide a theoretical basis, where we take a closer look at explanations and the characteristics of objects of explanations. Thereafter, we delve into mental representations of knowledge and interests before we segue to the research questions.
\section{Background}
\textbf{Explanations.} Explanations are an integral part of our daily lives and shape our understanding of the environment we live in \cite{ref_article24,ref_article27}. However, explanation generation is not yet fully explored and understood, as this process is intricate and encompasses diverse phenomena \cite{ref_article29}. Research reveals a tendency toward scientific explanations, whereas everyday explanations remain largely unstudied; the latter differ in structure and goals and function fundamentally differently from scientific explanations \cite{ref_article52}. Generally, people give explanations based on their knowledge, which is represented in the form of mental representations \cite{ref_article6,ref_article24}. In everyday explanations, the interlocutors play different roles. The explainer, EX, gives the explanation. The explainee, EE, receives the explanation regarding the object of explanation, which is called the explanandum \cite{ref_article17}. This older definition describes the EE as a mere passive explanation receiver.\\ 
From a co-constructive perspective on explanations, EXs and EEs actively develop the explanation conjointly. Co-constructive explanations represent a form of social interaction, where interactivity and partner relatedness are in the foreground. Both interlocutors interact with each other and negotiate which aspects of the explanandum need to be addressed throughout the explanation, as both interaction partners have their needs and aims regarding the explanation. A prerequisite for this reciprocal interaction is consideration of the other interlocutor \cite{ref_article39,ref_article46}. Thus, in co-constructive explanations, EEs and their explanation needs are taken more into account. Therefore, shifts in the course of explanation can occur as explanations are adapted toward implicit and explicit needs and toward developing understanding \cite{ref_article39}. As the explanandum can be examined and understood from different perspectives, the explanation content can shift depending on the selected perspective. Therefore, we follow a theoretical approach that is used for analyzing explanda from two distinct perspectives. This theoretical approach might also help to better understand the mental representations the EX has of the EE’s knowledge and interests regarding the explanandum in the context of artifacts like XAI. \\
\newline
\noindent\textbf{Technological Artifacts and the Dual Nature Theory.} In the context of XAI, the importance of technological artifacts must not be underestimated because XAI systems themselves are technological artifacts and, thus, need to be explained. The main characteristics of technological artifacts is that they are made by humans to fulfill certain purposes \cite{ref_article26,ref_article50}. According to the philosophy of technology, a prominent feature of technological artifacts is their dual nature—comprising an Architecture and a Relevance side \cite{ref_article26,ref_article44,ref_article49}. Architecture incorporates observable features that span from mechanisms to structure, which can be interpreted as an objective perspective on technological artifacts. In contrast, Relevance refers to goals, purposes, and intentions, thereby resembling a subjective perspective.\\ 
When explaining technological artifacts, both sides of the duality can be addressed \cite{ref_article26,ref_article50} and obviously both sides can be demanded by users of XAI. For example, users could ask for more information regarding physical properties, such as the code and underlying algorithms (Architecture) or about the functions and purpose of the system (Relevance). Therefore, the perspective of the dual nature theory might be suitable to analyze the knowledge and interests of EEs. Emphasizing one of the two dual sides during an explanation is not something that is done arbitrarily by any means. Indeed, both sides need to be considered in an explanation if the goal of the explanation is to achieve understanding \cite{ref_article43,ref_article47}. In a recent study, the structure of everyday explanations of technological artifacts in regard to the dual nature was investigated. The results indicated that explainers focused on physical components (Architecture) of the artifact first and then shifted toward interpretable components (Relevance). This was interpreted in such a manner that the components of the physical Architecture needed to be explained to create a foundation for the subsequent more complex aspects related to Relevance \cite{ref_article49}.\\ 
Further empirical research on the proposed dual nature theory in the context of everyday explanations, particularly on the mental representation crucial to explanations, is necessary. Therefore, in this study, we investigate the mental representations EXs have with regard to the interests and knowledge of EEs in terms of the dual nature of technological artifacts. According to the dual nature theory, both perspectives are necessary for EEs to completely understand the technological artifact. These mental representations are crucial to the explanatory process and, prospectively, XAI will hopefully be enabled to consider users and their needs in terms of the dual nature of technological artifacts. For this, XAI requires a component that monitors whether human end users require more information regarding aspects that can be allocated toward Architecture or Relevance. In naturalistic explanations, to effectively explain a technological artifact, the EX should be able to anticipate what the EE already knows regarding the technological artifact and which knowledge is still missing on both the Architecture and Relevance sides. This is because if the assumption regarding the EE’s interests and knowledge are not accurate, the explanation might not be satisfactory, as the focus could erroneously be on Architecture when Relevance is required or vice versa \cite{ref_article43,ref_article47}. Therefore, it is necessary for XAI to possess a mental representation of the end user that incorporates knowledge and interests. In the following account, aspects that constitute these mental representations are presented. \\
\newline
\noindent \textbf{The Partner Model.} Important aspects of evolving mental representations of interlocutors over the course of explanations fall within the scope of this paper. When communicating in the context of everyday explanations, interlocutors usually do not do this in isolation but in reference to a partner; hence, they behave in a partner-oriented manner \cite{ref_article3} and take each other into account \cite{ref_article2}. Being aware of the interlocutor is crucial to a suitable and satisfying explanation \cite{ref_article7}. The aforementioned awareness can be specified in terms of perceiving the interlocutors’ emotions and motivation as well as their prior and developing knowledge, interests, and ability to understand \cite{ref_article3,ref_article10}. The mental representation of these facets can be subsumed under the term “partner model”  \cite{ref_article10}, which is necessary for EXs to be able to react to EEs’ knowledge and interests \cite{ref_article2}. In other words, the explainer needs to know what the interlocutor already knows and wants to know \cite{ref_article3}. This is particularly important in terms to emphasize, as EXs can be biased toward their own knowledge and may not consider the needs of EEs to the fullest extent \cite{ref_article35}. The resulting misinterpreting needs of EEs is counterproductive \cite{ref_article51}. To avoid this problem, it is necessary that EXs accurately monitor EEs’ behavior and make assumptions regarding EEs \cite{ref_article7,ref_article8}. Among the diverse facets that constitute the partner model, emphasis is placed on interests \cite{ref_article20,ref_article25,ref_article45} and knowledge \cite{ref_article1,ref_article9}, as discussed in the following section.\\
\newline
\noindent \textbf{The Technical Model within the Partner Model.} The EXs’ partner models include information regarding the knowledge \cite{ref_article1,ref_article9} EEs have. As we are investigating everyday explanations of technological artifacts, EEs’ knowledge regarding technological artifacts is called a technical model. Therefore, we also speak of the technical model within the partner model. The technical model is represented mentally and is structured by the duality of technical artifacts. The technical model is required to be able to understand, to reason about, and foretell the states of technological artifacts \cite{ref_article18}. Within the technical model, different types of knowledge can be identified—for example, declarative knowledge \cite{ref_article32}, such as names and characteristics of elements, and procedural knowledge regarding the operation and manipulation of systems \cite{ref_article53}. Developing technical models are related to technical models of comparable systems \cite{ref_article40}. Because people have varying abilities and different interpretations of the purpose of technological artifacts, technical models are distinct and differ between individuals \cite{ref_article22,ref_article19}. In explanations, technical models develop over time. This process has been described as a multistage one \cite{ref_article16}. The initial stage is characterized by the identification of the elements a situation or system can possess, whereas the second stage is characterized by the identification of interrelatedness between the various elements and the ways in which interaction can occur \cite{ref_article32}.\\ When a technical model cannot be effectively utilized in specific contexts, technical models are updated by either adding novel and accurate information and/or by effacing erroneous information \cite{ref_article22,ref_article53}. In other words, the technical model is adjusted to permanently altering conditions and develops continuously through learning, experience, or interaction with domains \cite{ref_article19,ref_article22}. Differences in the structure of explanations have an impact on the developing technical model of EEs \cite{ref_article28}. Moreover, consideration of this technical model of EEs might enable the EX to give explanations that meet the needs of EEs. The following questions should help to illustrate what the technical model is embracing: “What does the interlocutor already know about a specific or similar domain in regard to the dual nature of technological artifacts?” or “Which knowledge does the interlocutor still need to acquire with regard to the dual nature of technological artifacts?” The technical model of the technological artifact is always perceived in terms of the dual nature of technological artifacts.\\
\newline
\noindent \textbf{Interests within the Partner Model.} The EXs’ partner models also include information regarding interests \cite{ref_article20,ref_article25}. When going back to the statement that an EX needs to know what the EE wants to know, one could construe this in the following manner: “What is the EE interested in?”. Interests play a pivotal role in explanations and have an impact on the direction an explanation takes. Furthermore, interests can be allocated to the two sides of the dual nature of technological artifacts. As EXs monitor EEs during an interaction, it is important to know how EXs build their assumptions regarding the EEs’ interests. This can be based on inferring in the situational context or by perceiving expressed interests through questions or by EEs’ comments \cite{ref_article45}. Here, interests refer to the technological artifact and its dual nature. It also beneficial for XAI to consider the interests of EEs, whether they are inferred or perceived and if they address Architecture or Relevance.  \\
\newline
\noindent \textbf{The Technical Model and Interests in the Context of XAI-Research.} Considering all these aspects, it becomes obvious that explanations should be customized to the needs of EEs in terms of the dual nature of technological artifacts. Ideally, XAI would be able to react to changes of varying degrees in knowledge and shifting interests during co-constructive explanations and address Architecture or Relevance when required. Therefore, it is necessary to understand how the technical model within the partner model develops. In this study, we approach the EXs’ assumptions regarding the EEs’ technical model and interests (see Fig. 1), which both are part of the partner model. We investigated what EXs assume and learn about EEs with regard to the dual nature of technological artifacts, namely Architecture and Relevance. When EXs consider aspects that they learn about EEs, it might be beneficial to the overall explanation. The EXs’ developing partner model might lead EXs through the explanation and to developing a better understanding of EEs \cite{ref_article35}. Because this also would hold true for explanations synthesized by XAI following this approach, the outcomes of this study are beneficial for XAI research and development. Then, working solutions toward the development of an advanced XAI that perceives and reacts to the knowledge and interests of EEs in a co-constructive manner could be realized in the future.\\
In our study, we investigated explanatory dialogues in which a simple technological artifact served as the explanandum—that is, the object of explanation. Further, we investigated how far EXs were aware of the knowledge and interests of EEs with regard to the dual nature of technological artifacts. The overarching question was that we addressed was: What does the EX think the EE needs in particular moments of an explanation? Thus, the focus lies on the perspective of EXs regarding EEs. We aimed to answer the following research questions: 
\begin{itemize}
    \item (RQ1) How do the EX’s assumptions regarding the EE’s technical models of the technological artifact with regard to the dual nature of technological artifacts develop during an explanation?
    \item (RQ2) How do the EX’s assumptions regarding the EEs’ interests in the technological artifact with regard to the dual nature of technological artifacts develop during an explanation?
\end{itemize}

\noindent
To answer these questions, we interviewed EXs to gain clarity on their assumptions of EEs’ knowledge and interests regarding the technological artifact. We first report the frequency of references allocated to Architecture or Relevance. Then we take an in-depth look at the content regarding EXs’ assumptions regarding the EEs’ technological model and interests and how they are related to the dual nature theory. Thereafter, EXs’ assumptions regarding the content of EEs’ technical model and interests are abstracted from the given context and technological artifacts to make it potentially useful for the XAI community.
\section{Method}
To investigate the technical model within the partner model, we conducted semi-structured interviews to assess the EX’s developing assumptions about the EE’s knowledge of and interests in the technological artifact in the course of an explanation. This was done with regard to the dual nature of technological artifacts in the context of naturalistic explanations. We designed a study in which we controlled specific variables, such as the location, explanandum (here we used the strategic game Quarto) and goal of the explanation. The study and procedure were standardized after pilot studies and the instruments were further developed and fine-tuned. The interviews were analyzed following a qualitative content analysis \cite{ref_article37} considering the two sides of the dual nature of technological artifacts: Architecture and Relevance as well as knowledge and interests.  \\
\newline
\noindent \textbf{Participants.}
Participants, EXs and EEs, were recruited for naturalistic explanations. The recruitment of participants followed a two-fold strategy: on-site recruitment and digital advertising. Communication with participants prior to the study ensured identical information for all participants. EXs were instructed to learn to play the game Quarto beforehand. This included familiarization with the game, playing the game, and explaining the game to others. EEs did not need to prepare for study participation. Both EXs and EEs needed to be proficient users of the German language, which is equivalent to C2 (CEFR). Permissions to conduct this study were granted by the Ethics Committee of Paderborn University and data protection was ensured.\\
We investigated nine explanations in dyadic settings EXs (\textit{N=9}). Participants were aged between 21 and 32 years (\textit{M=24.22, SD=3.46}). Three females and six males participated as EXs. All participants were students and have been enrolled, for example in education, linguistics, law, media studies, or computer science. Eight EXs were native speakers of the German language, while one EX was not a native speaker but had good command over the German language (C2). With regard to the experience with Quarto, EXs indicated that they played 0–18 rounds prior to the study (\textit{M=5.66, SD=6.14}). In addition to merely playing Quarto, four EXs stated that they explained the game to others. The number of explanations given ranged from one to three. Further, six EXs stated that they were experienced in explaining due to their side jobs. \\
\newline
\noindent \textbf{Procedure and Material.}
The technological artifact that served as our explanandum is the strategic board game called Quarto. The game is a two-player game, which requires logical reasoning. Quarto is deterministic in nature, providing all information to the two players. We considered it reasonable to use Quarto in this study, as games are, by definition, technological artifacts. Moreover, playing and, therefore, explaining games to others is a widespread social practice. In the explanatory process, the dual nature of games is expected to unfold as EX and EE need to address both Architecture (e.g., pieces, rules, goals) and Relevance (e.g., appliance of strategies, meaning of rules, and complexity). We decided to investigate a relatively simple game before delving into digital apps, which inherently have more depth.\\
The study consisted of three stages: pre-interview, explanation, post-interview with video recall-interviews. All semi-structured interviews were audio-recorded. Upon arrival, accidental encounters of EXs and EEs were avoided. EXs and EEs answered sociodemographic questionnaires. Thereafter, the semi structured pre-interview was conducted with the EXs. Part of the interview were questions that addressed the EEs’ technical model and interests: “What does the EE know about board games?” “Which aspects of board games does the EE consider as important?” and “Which aspects of board games does the EE enjoy?” The interview captured the EE’s technical model and interests with regard to Architecture and Relevance before the first encounter between EX and EE. Subsequently, the EX was asked to explain the game, “so that the EE could potentially win the game.” The EE was informed that a board game will be explained and was asked to participate actively. Thereafter, the explanation, in the absence of the game, began. It was video-recorded and monitored via livestream in order to enable researchers to follow the explanation in real time. After the explanation, a semi-structured post-interview was conducted to capture the EE’s technical model and interests after the explanation, which resembles a retrospect perspective. Apart from the same questions as in the pre-interviews, the following questions were included: “Which aspects of Quarto did the EE understand?” “Which aspects of Quarto did the EE consider important?” “Which aspects of Quarto did the EE enjoy?” Thereafter, video recall-interviews \cite{ref_article15,ref_article30} were conducted. The video recall-interview served as a method to reassess the EEs’ emerging and developing technical model and interests in concrete moments of the explanation. To be more specific, we utilized the video recall-interview to describe what the EX assumed about the EE during the developing explanation and how the assessed information relates to the two sides of the dual nature of technological artifacts. The following questions were asked: “What knowledge needs regarding the game did the EE have in that particular moment?” “What aspects of Quarto does the EE not know yet?” “How did the EE’s understanding of Quarto develop in this particular moment?” For the video recall, important explanation sequences were selected by utilizing predefined identification criteria, which included contributions of EX or EE that were interpreted as substantial, other- or self-initiated repair, turn-taking, misapprehensions, and motivational checks to assess understanding of the EE.\\ As phases of the explanation are characterized by varying participation of interlocutors—for example, the start is characterized as monological and the following phase as dialogical \cite{ref_article12}, we aimed at selecting two scenes each for the start, middle, and end of explanations. The time between the end of the explanation and the video recall-interview was as short as possible to assess the EE’s technical model before informational content could be transferred to long-term memory and to prevent a merger with conjoined memories and experiences \cite{ref_article31}. For video recall-interviews, the recorded explanation was shown to the EX and was stopped at the previously identified time marks. The selected scenes, as stimuli, enabled the EX to report elaborately regarding what they learned about EEs \cite{ref_article15}. Further, the selected stimuli allowed the EX to relive particular moments of the explanation, which helped to recall and interpret specific scenes \cite{ref_article30}. \\
\newline
\noindent \textbf{Content Analysis of Semi-Structured Interviews.}
All interviews were transcribed using standard orthography \cite{ref_article36}. On average, interviews (pre-, post- and video recall) of a single study lasted 58:15 mm:ss (\textit{SD=15.97}), ranging across all studies from 40:41 to 94:17 mm:ss. 
In total, over 08:44 hh:mm of interview material was transcribed and a sufficient amount of data was available for analysis. As interviews were rich in content, qualitative content analysis \cite{ref_article37}was opted. For transcribing and coding, MaxQDA software was utilized. 
The deductive coding manual resulted from an in-depth study of relevant literature work and analysis of pilot studies. The coding manual provided a short overview of the characteristics of knowledge, interests, Architecture and Relevance and included typical examples in the form of direct quotes (see Table 1). 
\begin{table}[h!]
\scriptsize
\caption{Code System with Typical Examples.}\label{tab1}
\begin{tabular}{l|l} \hline  
         \textbf{Categories}& \textbf{Typical Examples}\\ \hline  
         \multirow{1}{1.6cm}{\textbf{Knowledge}}  \\   
          {Quarto}& not considered \\
          \rowcolor{veryLightGray}
          {\hspace{0.2cm}{Architecture}}&By now she just knew the rules.(VP16,VR3,Pos.7)\\ 
          \rowcolor{veryLightGray}
         & That you have to build rows to win.( VP17,Post,Pos.21)\\ 
         {\hspace{0.2cm}{Relevance}} & She understood the game’s idea and everything behind it.(VP18,VR4,Pos.18) \\
         &She needed my experience on how to recognize situations.(VP26,VR5,Pos.17) \\
            {Board games}&  not considered\\  
         \rowcolor{veryLightGray}
           {\hspace{0.2cm}{Architecture}}& Everyone has in such a board game own pieces and colors.(VP24,Pre,Pos.38)\\
           \rowcolor{veryLightGray}
           & That you play in turns.(VP18,Pre,Pos.39)\\  
        {\hspace{0.2cm}{Relevance}} & He knows how to develop personal, cooperative strategies.(VP20,Pre,Pos.31) \\
       &Playing with several persons in a non-competitive way.(VP18,Pre,Pos.29)\\ \hline
                \textbf{Interests} &  \\   
         {Interests} & not considered\\
         {\hspace{0.2cm}{Expressed}}&not considered\\ 
      \rowcolor{veryLightGray}   
          Interests Architecture & Can you also build a diagonal row?(VP17,VR2,Pos.7)\\   
          \rowcolor{veryLightGray}
         &What does the game look like?(VP18,VR2,Pos.13)\\  
           {\hspace{0.2cm}{Expressed}}& Do I choose a piece for you? Or do I choose for myself?(VP26,VR3,Pos.5) \\
          &Is there exactly one pieces that is big, blue and round?(VP24,VR5,Pos.3)\\
\rowcolor{veryLightGray}
          Interests Relevance & That you learn something in a game. A strategy.(VP20,Pre,Pos.39) \\
          \rowcolor{veryLightGray}
           & That it is appealing in terms of color and visuals.(VP17,Pre,Pos.51)\\
         {\hspace{0.2cm}{Expressed}} & It is not about knowledge anymore but about experience.(VP26,VR5,Pos.9)\\
         & Why should I give you that piece so that you can win?(VP17,Post,Pos.11)\\
 \hline\end{tabular}
\end{table}\\
It was critical for the determination of segments—which could vary in length from single words to whole sentences—that one specific aspect addressed knowledge or interests \cite{ref_article42}. Thereafter, segments were coded with knowledge or interest categories addressing Architecture or Relevance. If segments could be coded to both Architecture and Relevance, both categories were assigned. If neither Architecture nor Relevance were addressed, the higher-ranking category, knowledge or interests, was selected. Codings from two coders were compared and discussed. Thereafter, the coding manual was revised in two iterations. Finally, all transcripts were coded by using the revised coding manual. Cohen’s Kappa was calculated to determine the intercoder reliability between two coders and it was k=.75 based on three coded studies. Values $\geq $.75 indicate excellent agreement \cite{ref_article13}.\\
Within the transcripts of the nine studies, a total of 991 segments were coded with Architecture and Relevance, as seen in Table 2. There were another 83 segments that could not be allocated to either Architecture or Relevance.
\begin{wraptable}{l}{0.55\textwidth}
\scriptsize
\captionsetup{singlelinecheck=false, justification=raggedright}
\caption{Allocation of Segments.}\label{tab1}
\begin{tabular}{l|l|l|l}
\hline
  & Architecture & Relevance & Total \\ \hline 
Knowledge & 353 (66.23\%) & 180 (33.77\%) & 533 (53.78\% \\
Interests & 259 (56.55\%) & 199 (43.45\%) & 458 (46.22\%) \\  \hline
Total & 612 (56.98\%) & 379 (35.25\%) & 991 (100\%) \\ \hline
\end{tabular}
\end{wraptable}
 These segments were not considered in further analyses. Usually, these segments were not rich in content—for example, when EXs could not answer questions. The number of coded segments does not necessarily translate to the number of aspects mentioned in interviews, as certain aspects were mentioned repeatedly and then were coded repeatedly as well. For the video recall-interviews, it was intended to identify two scenes for the start, middle, and ending of explanations. But as identification of scenes was complex and there were differences in the development of single explanations, we had 15 video recall interviews in the start, 24 in the middle, and 11 in the ending of explanations.
\section{Results}
We answered research questions in a twofold manner. We began with qualitative content analyses by transformation from transcribed interviews to the numbers of codes for each knowledge category.\\
\newline
\noindent\textbf{Code Frequency of Knowledge Categories.} To answer RQ1 (“How do EXs’assumptions regarding the EEs’ technical models of the technological artifact with regard to the dual nature of technological artifacts develop during the explanation?”), EXs’ mentions of assumptions regarding the EEs’ technical model in each category and explanation phase were counted (see Table 3).
In pre-interviews, the EXs’ assumptions regarding the EEs’ technical model of board games were mostly allocated to Architecture and to a lesser extent to Relevance. In the video recall-interviews in the start of explanations, EXs had limited assumptions regarding EEs’ technical model of board games in general. But EXs expressed beliefs regarding what EEs knew about Quarto. Again, coded segments were mainly referring to the Architecture and to a lesser extent to the Relevance. In video recall-interviews in the middle of explanations, EXs were almost exclu-
\begin{wraptable}{l}{0.7\textwidth} 
\scriptsize
\captionsetup{singlelinecheck=false, justification=raggedright}
\caption{Absolute Frequency of Knowledge Segments.}\label{tab1}
\begin{tabular}{l|l|l|l|l|l}
\hline
 & Pre &  VR-S & VR-M & VR-E & Post \\
\textbf{Knowledge} &  &  (15 VR) & (24 VR) & (11VR) & \\
\hline
\multirow{1}{2,5cm}{Board Games A} & 46 & 3 & 1 & 0 & 42 \\
\multirow{1}{2,5cm}{Board Games R} & 35 & 1 & 0 & 0 & 31 \\
\multirow{1}{2,5cm}{Quarto A} & 0 & 50 & 88 & 31 & 92 \\
\multirow{1}{2,5cm}{Quarto R} & 0 & 12 & 32 & 14 & 55 \\ \hline
Total & 81 & 66 & 121 & 45 & 220 \\ \hline
\multicolumn{6}{l}{Video Recalls of Explanation Phases; S:Start; M:Middle; E:End}
\end{tabular}
\end{wraptable}
sively referring to EEs’ technical models in regard to Quarto. In most cases the segments addressed Architecture. Further information was primarily related to prior knowledge. Segments addressing Relevance increased and accounted for over a quarter of the coded segments. In video recall-interviews, at the end of explanations, EXs had strong beliefs regarding the EEs’ technical models regarding Quarto, and the majority of segments were coded as Architecture. The share of segments addressing Relevance peaked, with almost a third of all coded knowledge segments. In post-interviews, EXs described their assumptions primarily regarding EEs' technical model of Quarto. It became evident in EXs’ assumptions that EEs’ technical model referred to both Architecture and Relevance. In the following account, concrete aspects of the technical model are presented.\\
\newline
\noindent \textbf{Content of Knowledge Categories.} After presenting the frequency of coded knowledge segments, we now shed light on which concrete aspects of technological artifacts EXs referred to when they spoke about EE’s technical model. We bundled statements of EXs on EEs’ technical model (see Table 4). The aim is to present aspects of this particular technological artifact EXs believed EEs had knowledge about.\\
In \textbf{pre-interviews} coded segments were usually answers to the interview question “What does the EE know about board games?” EXs assumed what the, still unknown, EEs might know about board games. Hence, the EXs’ assumptions regarding EEs’ technical model were often influenced by common and individual knowledge EXs themselves had regarding games. On the Architecture side, EXs reported that EEs might know that board games usually have goals, are turn based, and include pieces. On the Relevance side, EXs mentioned that EEs might know that board games are typically played in a social context, that games are for teaching as well as learning, and that EEs knew how to develop and deploy strategies. From a general perspective on board games, incorporating both duality sides, EXs stated that EEs might know diverse game genres and mentioned concrete examples. Even before the start of explanations, EXs appeared to naturally have relatively clear assumptions regarding EEs’ technical models, which then were tested by EXs over the course of explanations, which led to a revised technical model.  \\
In the video recall-interviews EXs were asked these questions: “What aspects of Quarto does the EE not know yet?” and “How did the EE’s understanding of Quarto develop in this particular moment?” In video recall-interviews regarding \textbf{the start of explanations}, EXs’ beliefs regarding EEs’ technical models were partially based on assumptions that EXs already had prior to explanations, interference from the given explanation and partially from interaction with EEs. Here, EXs stated that EEs knew, to a certain extent, the physical components of the game, its specific rules, and the goal. After inspection of explanations, we knew that these assumptions aligned relatively well with the content of explanations. EXs also spoke about missing knowledge of EEs. This referred particularly
\begin{table}[h!]
\scriptsize
\caption{Development of Knowledge over the Course of Explanations.}\label{tab1}
\begin{tabular}{ll|l|l|l|l|l|l|l|l}
\hline
 &   & \multicolumn{8}{c}{\textbf{Interviews}}  \\
 & & \textbf{Pre} &\multicolumn{2}{|l|}{\textbf{VR-Start}} &\multicolumn{2}{|l|}{\textbf{VR-Middle}}& \multicolumn{2}{|l|}{\textbf{VR-End}} & \textbf{Post}\\ 
\hline
\rowcolor{veryLightGray}
\multicolumn{2}{l|}{\textbf{Knowledge aspects}}  & PK & GQK  & DQK & GQK & DQK & GQK & DQK  & K  \\ \hline
{\textbf{Architecture}} & & & & & & & & \\
\rowcolor{veryLightGray}
\underline{Overall game} & &&&&&&&& \\
{\hspace{0.1cm}{Goal of game}}  & & 18,24 & 17,18 &  & 17,18 & 17,18 &  &  & \\
\rowcolor{veryLightGray}
{\hspace{0.1cm}{Look of game}}  &  & &  & 16,21 &  & 17,20,25 &  & 16,17 & 17,20\\
\rowcolor{veryLightGray}
 &   &  &  &  &  & 26 &  & 24 & 25,26 \\
\underline{Material components} & &&&&&&& \\
\rowcolor{veryLightGray}
 {\hspace{0.1cm}{Pieces}}  & & 17,21,24 & 18,20 & 21,24 &  &  &  &  & 18,21\\
 \rowcolor{veryLightGray}
   & &  & & 25 &  &  &  &  & 26\\
 {\hspace{0.1cm}{Utilization of pieces}}  & &  & 17,18,20 &  & &  &  &  &  \\
   & & & 21,25 &  & &  &  &  &  \\
\rowcolor{veryLightGray}
\underline{Immaterial components} & &&&&&&&& \\
 {\hspace{0.1cm}{Course of the game}}  &  & & & 20,25 &  & &  &  & \\
 \rowcolor{veryLightGray}
 {\hspace{0.1cm}{Turns}}  & & &  &  & 17,25,26 & 17,25,26 & 20,21,24 & 20,21,24 &  \\
 {\hspace{0.1cm}{Rules}} &  &  & 16,17 & 24,25 & 16,17,18 & 16,17 & 16,21 & 16,21 & 18,21 \\
 &  &  &  &  & 24,25 & 24,25 & 24 &  24& 26\\
 \rowcolor{veryLightGray}
{\hspace{0.1cm}{Prerequisites for}}  &  &  &  &  &  &18,20  &  &18,20 & \\ 
\rowcolor{veryLightGray}
{\hspace{0.1cm}{winning}}  &  &  &  &  &  &24  &  & & \\  \hline
{\textbf{Relevance}} & &&&&&&& \\
\rowcolor{veryLightGray}
\underline{Overall game} & &&&&&&&& \\
 {\hspace{0.1cm}{Complexity of the game}}   & &  &&  & 24 &  & 17 &  & \\
 \rowcolor{veryLightGray}
 {\hspace{0.1cm}{Ideas behind the game}}  &  &  &  && 18 &  &  &  & \\
\underline{Immaterial components:} &&&&&&&& \\
\rowcolor{veryLightGray}
 {\hspace{0.1cm}{(Relevance of) rules}}  &  & & & 24 & 17,18 &  & &  & \\
 {\hspace{0.1cm}{(Deploying) strategies}}  &  &20 &  &  & 17 & 17,18,25 &20,25,26  & 20,25,26 & 17,24,26   \\
 \rowcolor{veryLightGray}
 {\hspace{0.1cm}{Challenging aspects }}  &&  &  &  & 17,21,22 &  & & & \\
 \rowcolor{veryLightGray}
   &&  &  &  & 24,26 &  & & & \\
\underline{Purpose of board games:}  & & & & && & & &17,24,26  \\
  \rowcolor{veryLightGray}
 {\hspace{0.1cm}{Teaching and learning}}  & & 18 &  &  &  &  &  &  &  \\
 {\hspace{0.1cm}{Game's context of usage}} &  & 18 &  &  &  &  & &  &  \\ \hline
{\textbf{Architecture/Relevance}} & &&&&&&&\\
\rowcolor{veryLightGray}
 {\hspace{0.1cm}{Board games:}}  & & 16,18 &  &  &  &  &  &  &  \\
\rowcolor{veryLightGray}
& & 21,24,25 &  &  &  &  &  &  &  \\
{\hspace{0.1cm}{Game genres:}} &  & 17,18,26 &  &  &  &  &  &  & 17,18 \\
  &   &   &  &  &  &  &  &  & 21,22 \\
\rowcolor{veryLightGray}
 {\hspace{0.1cm}{Concrete examples:}}  & & 16,18,25 &  &  &  &  &  &  & 16,17\\
 \rowcolor{veryLightGray} 
  &   & 26 &  &  &  &  &  &  & 18,22 \\ \hline
  \multicolumn{10}{l}{PK: Prior Knowledge, GQK: Gained Quarto Knowledge,DQK: Deficits Quarto Knowledge}\\
\multicolumn{10}{l}{K: Knowledge. VR: Video Recall. Numbers in table refer to study number: VP}\\
\end{tabular}
\end{table} \\
to Architecture knowledge, which was not covered yet in explanations—for example, clear ideas of how the game looks, details regarding the characteristics of pieces, the overall course of the game, and complex rules. Accordingly, the technical model also contains information regarding missing knowledge.\\
With regard to the \textbf{middle of explanations}, a shift from a predominantly monological to dialogical interaction occurred. In contrast to the start of explanation, EXs now had developed strong beliefs regarding the EEs’ Relevance knowledge of Quarto. This was due to the fact that EXs perceived EEs’ signals of understanding—for example, through questions, statements, or multi-modal behavior. On the Relevance side, EXs believed EEs knew more about challenging aspects of the game, its interesting aspects, and the complexity of the game. EXs also had strong beliefs regarding the Architecture knowledge and usually believed that EEs did know numerous characteristics of game components, material or immaterial. Nevertheless, EXs stated that EEs did not have complete knowledge at this point. They missed knowledge regarding the look of the game, rules, game turns, and prerequisites for winning on the Architecture side. With regard to Relevance, EXs reported that EEs might need insights on personal experience within certain situations—or examples, when deploying strategies. In conclusion, for this phase, it can be said that EXs tracked the explanation content and clearly monitored consolidated and missing knowledge.\\
Toward the \textbf{end of explanations,} EXs beliefs shifted from being based mainly on assumptions to what they learned through EEs’ questions, statements, and summaries. With regard to the end of explanations, no significant changes in assumptions regarding EEs’ technical models were described. EXs focused on aspects that were understood. EXs concluded that EEs understood most aspects on both the Architecture and Relevance sides. However, EXs believed that there were knowledge gaps and details that remained unclear. However, this missing knowledge was not important for a potential application of knowledge. Therefore, these aspects were not explained in a closing manner. This referred mainly, on the Architecture side, to a clear visual imagination of the game. In contrast, with regard to Relevance, EXs reported insufficient and/or missing knowledge regarding gameplay experience, such as knowing how to behave in certain game situations. This included information regarding what to consider and what to do how in which order. Here, the interrelatedness of explanation content became more prominent. EXs considered these aspects to be important for EEs but partially had the impression that EEs did not completely understand. This might be an indication for aspects that require more than an abstract explanation and that knowledge needs to be tested or experimented with. \\
To summarize, it can be said that at the end of the explanations, EEs did not understand all aspects of the game equally well. In certain cases, something was missing or could not be entirely resolved. But EXs generally developed well-defined mental representations of EEs’ technical models—for example, they had concrete beliefs regarding what EEs knew and did not know. Despite the fact that not every single aspect was understood to the furthest possible extent, EXs generally had the feeling that no further explanation was required due to the fact that a) not all aspects are equally important and b) not all aspects can be explained completely satisfactorily in an abstract manner.\\
In \textbf{post-interviews}, referring to the EEs’ technical models after the explanation, the same questions as that in the pre-interviews were asked. In addition, the following question was added: “Which aspects of Quarto did the EE understand?” EXs were referring to prior knowledge of EEs and again mentioned types of games and provided examples. In particular, EXs were referring to games that helped EEs to understand aspects of Quarto on both the Architecture side (from the game components to rules) and the Relevance side (from the purpose of games to strategies). When EXs described their beliefs regarding EEs’ technical model of Quarto in post-interviews, the majority had the impression that EEs understood the game relatively well—for example, the various Architecture aspects. With regard to Relevance knowledge, EXs mainly referred to strategic elements and degree of complexity. But it was also emphasized that EEs still had to overcome knowledge gaps or lack of clarity—for example, regarding the visuals, characteristics of game components (Architecture), and details regarding implementing strategies (Relevance). EXs speculated that EEs probably needed to apply the gained knowledge by interacting with and playing the game in order to overcome these deficits in knowledge. Within the interviews, it became obvious that EXs started with assumptions regarding Architecture and Relevance knowledge and that these assumptions, at least partially, grew into certainty following the course of the explanation. \\
\newline
\noindent \textbf{Code Frequency of Interest Categories.} 
\noindent To answer RQ2 (How do EXs’ assumptions regarding the EEs’ interests in the technological artifact regarding the dual nature of technological artifacts develop during the explanation?), EXs’ mentions of assumptions regarding the EEs’ interests in the technological artifact in each category and explanation phase were counted (see Table 5). 
\begin{wraptable}[12]{r}{0.68\textwidth} 
\scriptsize
\captionsetup{singlelinecheck=false, justification=raggedright}
\caption{Absolute Frequency of Interests Segments.}\label{tab1}
\scriptsize 
\begin{tabular}{l|l|l|l|l|l} \hline
 & Pre &  VR-S & VR-M & VR-E & Post \\
\textbf{Interest} &  &  (15 VR) & (24 VR) & (11VR) & \\
\hline
\multirow{1}{3,5cm}{Interest A} & 58 & 15 & 16 & 2 & 49 \\
Expressed Interest A & 0 & 10 & 50 & 25 & 34 \\
Interest R & 107 & 1 & 5 & 0 & 45 \\
Expressed Interest R & 0 & 1 & 18 & 2 & 20 \\ \hline
Total & 165 & 27 & 89 & 29 & 148 \\ \hline
\multicolumn{6}{l}{Video Recalls of Explanation Phases; S:Start; M:Middle; E:End}
\end{tabular}
\end{wraptable}

\noindent In pre-interviews, there were almost twice as many segments regarding Relevance interests compared to Architecture interests. In contrast, in video recall-interviews in the start of explanations, EXs believed EEs were particularly interested in aspects addressing Architecture and also referred to directly expressed interests. In video recall-interviews in the middle of explanations, EXs perceived EEs’ increased interests in Relevance aspects. The majority of perceived interests were directly expressed. In video recall-interviews at the end of explanations, EXs mentioned that EEs strongly expressed interests in Architecture.
In post-interviews, EXs mentioned assumed and expressed interests in game aspects with regard to both sides of the duality. In the following account, concrete aspects of these interests are presented.\\
\\
\\
\noindent \textbf{Content of Interest Categories.} In the following section, concrete content in EXs assumptions regarding EEs’ interests in the technological artifact (here Quarto) and other comparable artifacts (other games) are presented. Statements on EEs’ interests were bundled as well (see Table 6.). In \textbf{pre-interviews}, EXs were asked which aspects of games EEs might be interested in. With regard to Architecture, the EXs’ assumptions were particularly related to EEs’ interests in immaterial components of the game, such as rules, the goal, and the concept of the game. With regard to the Relevance, EXs assumptions regarding EEs’ interests were manifold, ranging from an appealing design, 
\begin{table}[h!]
\scriptsize
\caption{Development of Interests over the Course of Explanations.}\label{tab1}
\begin{tabular}{ll|l|l|l|l|l|l|l|l}
\hline
 &   & \multicolumn{8}{c}{\textbf{Interviews}}  \\
 & & \textbf{Pre} &\multicolumn{2}{|l|}{\textbf{VR-Start}} &\multicolumn{2}{|l|}{\textbf{VR-Middle}}& \multicolumn{2}{|l|}{\textbf{VR-End}} & \textbf{Post}\\ 
\hline
\rowcolor{veryLightGray}
\multicolumn{2}{l|}{\textbf{Interests aspects}}  & AI & AI  & EI & AI & EI & AI & EI  & I  \\ \hline
{\textbf{Architecture}} & & & & & & & & \\
\rowcolor{veryLightGray}
\underline{Overall game} & &&&&&&&& \\
{\hspace{0.1cm}{Goal of game}}  & & 18,25 &  &  &  & 18,21,24,25 &  &  & \\
  \rowcolor{veryLightGray}
{\hspace{0.1cm}{Multiplayer concept}}  &  & 16 &  &  &  &  &  &  & \\
{\hspace{0.1cm}{Game´s length}}  &  & 16,18 &  &  &  &  &  &  & \\
\rowcolor{veryLightGray}
{\hspace{0.1cm}{Look of game}} &  & &  &  & 18,25 &  & 17  &  & 20,21,24\\
\underline{Material components}  &  & 17,21,22,26 &  &  &  &  &  &  & \\
 \rowcolor{veryLightGray}
 {\hspace{0.1cm}{Pieces}}  &  & &18,25  &  &  &  &  & 16,18,21 & \\
 \rowcolor{veryLightGray}
 &  &  &26  &  &  &  &  & 22,25 & \\
  {\hspace{0.1cm}{Characteristics of pieces}} &  &  & 20,24 & 17,20 & 17,25 &  & & 20,21,24 & 17,20 \\
  &  &  &26  &  &  &  &  &25  & 22,24 \\
 \rowcolor{veryLightGray}
 {\hspace{0.1cm}{Utilization of pieces}} &  &  & 21,25 &  & 16,17,25 &  & & 21 & 17,25\\
\underline{Immaterial components} & &&&&&&& \\
 \rowcolor{veryLightGray}
 {\hspace{0.1cm}{Course of the game}}  &  & & &  & 18,20  & &  &  & \\
 {\hspace{0.1cm}{Turns}}  &  & 21,25,26 &  &  &  &  &  &  & 25,26\\
  \rowcolor{veryLightGray}  
 {\hspace{0.1cm}{Rules}} &  & 16,18,21,22 &  &  &  & 17,20,21,22 &  & & 16,17,18 \\
 \rowcolor{veryLightGray}
&  & 25,26&  &  &  & 25,26 &  & & 20,24 \\
\hline
{\textbf{Relevance}} & &&&&&&& \\
 \rowcolor{veryLightGray}
\underline{Overall game} & &&&&&&&& \\
 {\hspace{0.1cm}{Complexity of the game}}   & & 16,17,18,20  && 24 &  & 24 & &  & \\
    & & 21  &&  &  &  & &  & \\
     \rowcolor{veryLightGray}
 {\hspace{0.1cm}{Special feature of the game}}  &  &  &  &&  & 18 &  &  & 22,26\\
 {\hspace{0.1cm}{Appealing design}}  &  &  17 & &&  &  &  &  & \\
  \rowcolor{veryLightGray}
 {\hspace{0.1cm}{Unpredictability in  }}  &  & 16 & &&  &  &  &  &\\
  \rowcolor{veryLightGray}
  {\hspace{0.2cm}{course of the game}}  &  &  & &&  &  &  &  &\\
\underline{Immaterial components:} &&&&&&&& \\
  \rowcolor{veryLightGray}
 {\hspace{0.1cm}{Meaning behind rules}}  &  &  & &&  & 17 &  &  & \\
 {\hspace{0.1cm}{(Deploying) strategies}}   &  &  18,20,21,22 & &&  & 20,22 &  & 25,26  &21,22,24 \\
     &  &  24,26 & &&  &  &  &   &25,26 \\
 \rowcolor{veryLightGray}    
\underline{Purpose of board games:}  & & & & && & & &17,24,26  \\
 {\hspace{0.1cm}{Immersion in games}}  & & 18,24,25,26 &  &  &  &  &  &  &  \\
  \rowcolor{veryLightGray}
  {\hspace{0.1cm}{Winning}} &  & 17,18,20,21 &  &  &  &  & &  &  \\ 
  \rowcolor{veryLightGray}
  &  & 22,24,25,26 &  &  &  &  & &  &  \\ 
 {\hspace{0.1cm}{Teaching and learning}} &  & 18,20,22,24 &  &  &  &  & &  &  \\ 
  \rowcolor{veryLightGray}
 {\hspace{0.1cm}
 {Opportunity for socializing}} &  & 18,20,21,22 &  &  &  &  & &  &  \\ 
    \rowcolor{veryLightGray}
    &  & 24 &  &  &  &  & &  &  \\ \hline
  \multicolumn{10}{l}{AI: Assumed Interests , EI: Expressed Interests, I: Interests.}\\
  \multicolumn{10}{l}{VR: Video Recall. Numbers in table refer to study number: VP}
\end{tabular}
\end{table} 
challenges, and complexity of the game to strategies. EXs mentioned their assumptions regarding interests in the overarching purpose of games numerous times. They believed EEs were interested in games that offer opportunities to socialize, to learn and teach, to enjoy the immersion in games, and the winning experience. Thus, the overarching purpose of an artifact was emphasized and this probably could have marked a good starting point for the concrete explanation.  \\  
In the retrospective video recall-interviews regarding the interests, EXs were asked the following questions: “What knowledge needs regarding the game did the EE have in that particular moment?” Pilot studies indicated that EXs elaborated more on EEs’ interests when the phrase “knowledge needs” was used. This was probably due to the fact that EEs linked the question regarding interests to game aspects that people are generally interested in (see section regarding pre-interviews). In the context of video recall-interviews, this question was found to cause unelaborated and short answers, when particular explanation phases for concrete aspects of the game were provided. These rather specific aspects would probably not be mentioned if people generally were asked about their interests in games. Therefore, we decided to use the phrase “knowledge needs” to obtain richer answers regarding the interests.\\
In video recall-interviews regarding the \textbf{start of explanations}, in contrast to pre-interviews, EXs almost exclusively assumed EEs’ interests addressing Architecture, such as physical game components like pieces and their characteristics. As the explanation was just beginning, EXs generally only had vague assumptions regarding EEs’ interests and typically explained in a monological manner. In certain cases though, EXs reported that EEs already expressed interests in the start of explanations.  \\
After watching sequences from the \textbf{middle of explanations}, which were characterized by an increasingly dialogical interaction, EXs mostly mentioned expressed interests of EEs in the form of questions and statements. In comparison to expressed interests, EXs referred to assumed interests less frequently. In regard to Architecture, EXs had the assumptions that EEs would be interested in the look and course of the game as well as the characteristics of pieces. EXs also perceived expressed Architecture interests of EEs with regard to the goal and rules. Interestingly, EXs did not have exclusive assumptions regarding EEs’ interests in Relevance. Here, assumed Relevance interests were double coded, when game aspects were referred to from a global perspective. It needs to be highlighted that EXs reported on directly expressed Relevance interests of EEs, such as interests in the complexity of the game, strategies, special features, meaning behind rules, or insight on personal experience in concrete game situations.  \\
Not all interests that were assumed by EXs were actually expressed by EEs. More severe though is the fact that EXs struggled to anticipate the EEs’ interests in Relevance. Interests in Relevance were then directly expressed by EEs. For the explanation process, accurate assumptions regarding EEs’ interests are important. The EEs’ demands EXs reported on were particularly important for understanding and prospectively using the artifact. Thus, it is necessary that EXs adapt to the interests in the course of the explanations.\\ 
With regard to the \textbf{end of explanations}, EXs reported that almost all interests of EEs were directly expressed. Most EXs stated that EEs had interests in aspects attributed to pieces in one form or another. With regard to Relevance, EXs made no assumptions on interest but sporadically perceived expressed interests, such as aspects that were important to consider when deploying strategies and insights on gameplay experience. The number of times that EXs mentioned EEs’ interests decreased slightly. At the end of explanations, when numerous aspects were explained, interests of EEs were found to be abating. Therefore, ultimately, interests were less prominent but were still important, as EEs demanded answers for their last questions.  \\
In comparison with pre-interviews, where EXs developed numerous strong and diverse assumptions regarding EEs’ interests—probably due to the fact that EXs required hints for orientation in the explanation—in retrospect, \textbf{post-interviews} EXs summed up what they learned about EEs’ central interests during the explanation. No new aspects were introduced here. On the Architecture side, EXs reported that EEs were interested in rules, pieces, prerequisites for winning, and how the game looks. EXs also summed up the expressed interests. On the Relevance side, EXs mentioned that EEs were interested in special features of the game, strategies, and insights on gameplay experience and again summed up the expressed interests of EEs. Further, EXs considered increasingly expressed interests but also continued to infer interests.
\section{Discussion}
To be able to implement features of co-constructiveness in XAI it is necessary to learn more about naturalistic explanations among humans to fully understand how the EXs assumptions about the EEs’ knowledge and interests develop. The importance of the dual nature theory for comprehensible explanations has been previously highlighted \cite{ref_article49}. Thus, the focus of this paper was the investigation of assumptions of the EX regarding the knowledge and interests of the EE, with the introduction of the novel concept of technical models and interests within partner models with regard to the dual nature of technological artifacts \cite{ref_article26}. How they developed throughout the explanation, from a vague and initial to a well-defined technical model and interests within the partner model after explanations, was empirically shown in this paper. \\
\newline
\noindent\textbf{Answer to RQ1.} (“How do EX’s assumptions regarding the EEs’ technical models of the technological artifact in regard to the dual nature of technological artifacts develop during the explanation?”) We investigated which assumptions EXs had about EEs' technical model before, during, and after the explanation. The EXs spoke about assumptions they had about EEs’ knowledge. In pre-interviews, it was remarkable how many assumptions, also concerning Relevance, EXs had about the hypothetical knowledge of EEs. These early assumptions of EXs were tested throughout the interaction and gradually added to a refined technical model within the partner model \cite{ref_article22,ref_article53}.\\
Over the course of explanations, EXs’ assumptions regarding the EEs’ technical model developed. In the beginning of explanations, EXs particularly had assumptions regarding the EEs’ Architecture knowledge. Further information was added to the technical model within the partner model, as the EXs also made assumptions regarding EEs’ missing knowledge. This occurred throughout the continuing explanation. The EXs’ assumptions regarding the EE’s Relevance knowledge increased in the middle of explanations. The reason for this could be the increase in the number of expressed interests \cite{ref_article45} of EEs in the dialogical phase of explanations \cite{ref_article12}. The EXs were monitoring \cite{ref_article6,ref_article8} and reacting to these interests in further explanations and, therefore, built assumptions regarding the EEs’ developing Relevance knowledge. During the explanation, EXs seldom made assumptions regarding EEs’ knowledge about comparable technological artifacts—that is, prior knowledge. But occasionally this happened when EXs perceived that EEs made comparisons that were helpful for them to understand the technological artifact \cite{ref_article40}. Towards the end of the explanations, almost half of the EXs had assumptions regarding the EEs’ knowledge concerning Relevance. This was due to the fact that EXs increasingly perceived expressed interests of EEs in the middle of the explanation and then were referring to EEs’ developing Relevance knowledge at the end of explanations. It is noteworthy that EXs still perceived lacks in knowledge of EEs required for a potential usage of the explained technological artifact. Many EXs weren’t aware of this in the explanation itself but realized the knowledge gaps, particularly in the video recall-interview. The statements of EXs mainly referred to gaps in procedural knowledge \cite{ref_article1,ref_article9,ref_article53}, where interrelatedness of knowledge aspects needed to be made clearer \cite{ref_article32}. Additionally, EXs stated that EEs did not have a clear visual image of the technological artifact. Despite the lack of knowledge, a few EXs completed their explanations.\\ For XAI, this might be an implication that close monitoring is crucial. If XAI is aware of knowledge gaps, there still might be reasons for stopping the explanation. Because not every aspect needs to be completely understood, considering the potential context and extent of application. Alternatively, it might be beneficial to allow the interaction with the technological artifact to consolidate knowledge and practical application. The EEs’ earliest technical models were based on assumptions and could be described as a part of the global partner model. Throughout the explanation, EXs learned more about the EEs’ technical models through testing assumptions and monitoring the development of EE’s knowledge. The technical models were refined continuously, in a stepwise manner, until the end of explanations. \\
\newline
\noindent \textbf{Answer to RQ2.}(“How do EXs’ assumptions regarding the EEs’ interests in the technological artifact with regard to the dual nature of technological artifacts develop during the explanation?”) We investigated which assumptions EXs had about EEs interests in the technological artifact before, during, and after the explanation. It was evident that EXs in pre-interviews had rather strong assumptions about EEs’ interests in board games, particularly on the Relevance side and noticeably fewer assumptions on an Architecture side. This supports the idea of a well-defined partner model in that matter. EXs emphasized that EEs would be particularly interested in the overall purpose of the technological artifact. One could argue that this would be an ideal starting point for an explanation. But this contradicts with what happened in the starting explanation, where the focus was set on the Architecture of particular components of the technological artifact. This probably could be due to the fact that presumably EXs assumed that EEs knew what games are for and, therefore, did not need explanations. Thus, further research certainly is necessary to ascertain what happens in the start of explanations of novel technological artifacts that are set to fulfill specific and intended functions.\\
In the start of explanations, EXs almost exclusively had vague assumptions regarding EEs’ interests addressing Architecture. In the middle of explanations, after the shift from a predominantly monological to dialogical explanation phase \cite{ref_article12}, EXs learned more about EEs and their interests. Thus, interests were increasingly expressed \cite{ref_article45}. It is worth mentioning that EXs were not anticipating all interests of EEs and this was particularly true for Relevance aspects. We only can suspect the concrete reason underlying this. However, human explainers can only handle a certain amount of cognitive load \cite{ref_article48} and EXs mentioned repeatedly that they were not always able to focus on the EEs’ needs as they were more concentrated on providing the explanation. As EEs were expressing their interests, EXs were able to react and adapt to those interests \cite{ref_article2}. This reciprocal interaction reveals that both EX and EE negotiated the further direction of explanation \cite{ref_article46}. If EXs would not have been aware of EEs interests, the ongoing explanation probably would have been less satisfactory for EEs \cite{ref_article51}. The partner models concerning interests were further refined. Toward the end of the explanation, EXs reported that EEs were showing interests primarily in Architecture aspects, where further information or clarification was demanded. This indicates that EEs required explanation from both duality sides for understanding \cite{ref_article43,ref_article47} and a well-developed technical model for potential application of knowledge \cite{ref_article18}. Therefore, perspectives on the technological artifact were altering at different points in time and adapting to the needs of EEs. This process is already known from an observation study: After the beginning of explanations, duality perspectives of explanation content were permanently altered \cite{ref_article49}. Here, we investigated pre-selected scenes within explanations, but it became evident as well that both perspectives were required.\\
In post-interviews, distribution between EXs beliefs regarding EEs was relatively balanced with regard to the dual nature. The total number of assumptions in interest were significantly lower than those in post-interviews. This could probably be explained by the fact that explanations ended and interests did not matter as much as in the beginning, when interests ultimately had an impact on the directions that explanations took.\\
\newline
\noindent\textbf{Implications for XAI.} We consider the mental representation of the EEs’ knowledge and interests to be a prerequisite for satisfying explanations. After the initial phase of explanations—that is the monological phase \cite{ref_article12} in which basic structures of the technological artifact were explained \cite{ref_article49}, the EX needed to know where and how to continue the explanation and when to refine. Therefore, having an accurate representation of what the EE is interested in and which knowledge can be built on is absolutely mandatory in explanatory processes. Having a component in XAI that contains information regarding EEs’ knowledge and interests could enable those systems to be explained according to the needs of users. Close monitoring \cite{ref_article6,ref_article8} of EEs through the consideration of multimodal behavior, questions and statements would be required. Aspects that require consideration would, for example, be developing knowledge (which knowledge does the EE have and which knowledge is missing?) and interests (which interests does the EE have in the technological artifact?). Ideally, XAI will not only possess a well-conceptualized strategy in explaining concrete technological artifacts but also considers EEs and their needs \cite{ref_article23}, as both aspects are immensely important to satisfying explanations and higher levels of understanding: For example, regarding games, explaining strategies does not help when information on tricks to play them well are missing. EEs particularly needed a back and forth between perspectives in a) aspects of higher complexity b) novel aspects c) unexpected aspects. With regard to the dual nature of technological artifacts, both perspectives—Architecture and Relevance—were important for EEs to understand: a) EXs perceived different needs of EEs at different points in time regarding the artifact, and b) each artifact has its own unique features that might be challenging for EEs to understand. \\
\noindent After abstraction of content about knowledge and interests, indicators for synthesized explanations considering the dual nature of technological artifacts were derived (see Fig. 1). 
For example, a starting point for an explanation could be marked by high interests and low knowledge \cite{ref_article4}. 
\begin{figure}
\centering
\includegraphics[width=1\linewidth]{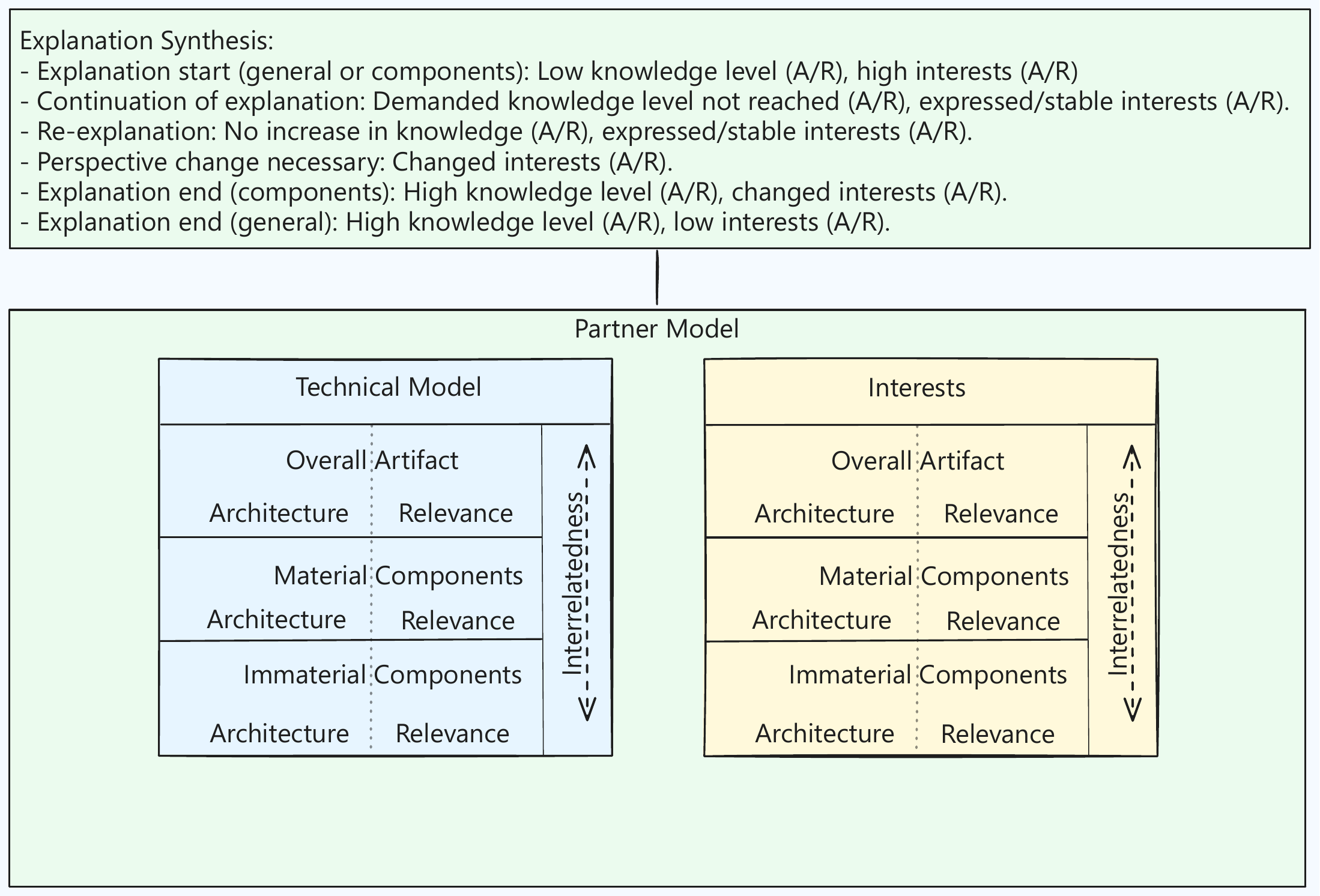}
\caption{The Partner Model for Synthesis of Explanations}
\label{fig:enter-label}
\end{figure}
It could be an indicator for continuation of the explanation when the requisite level of knowledge is not yet reached or interests regarding that aspect were expressed. When interests change, the foci or perspective of explanation might need to be changed as well. Moreover, a re-explanation of certain aspects might be necessary when interests do not change and knowledge does not increase further. Explanation could potentially be stopped when no interests are expressed and satisfying levels of knowledge are developed. With this background, we can deliver ideas for practical implications supporting a socio-technological approach for XAI \cite{ref_article34,ref_article41}: As a technical artifact, XAI already possesses the dual nature naturally. Therefore, we argue that the dual nature regarding knowledge and interests should be part of the XAI’s representations of end users to generate satisfactory explanations that meet users’ needs on both the Architecture and Relevance sides. With regard to this, XAI would require the technical and interests within the partner model—that is, a component with the very same function and capabilities. The diverse aspects of the user \cite{ref_article38} could then be considered by XAI. With a view to XAI that could potentially monitor and track \cite{ref_article6,ref_article8} the development of technical models and interests, the absence of mental load in explanations—which humans are affected by \cite{ref_article48}—could potentially enhance explanations synthesized by XAI. Moreover, XAI would potentially be able to be co-constructive from the start of explanations and, therefore, consider and react to the needs of EEs initially. This might be beneficial when only explanations on certain aspects of more complex technological artifacts are demanded by EEs.\\
\noindent \textbf{Methodological Considerations and Limitations.} There are distinct quality criteria for qualitative research \cite{ref_article37,ref_article14}. Hence, we followed advice and ensured that we meet quality standards in multiple ways. We selected research methods that were appropriate for research questions and the object of investigation. Instruments were carefully developed in consideration of thematic and methodological requirements. We ensured inter-subject comprehensibility of our study by documenting the research process (e.g., collection method, transcription rules, methods of analysis, and decisions regarding problems) and by discussing and interpreting data conjointly. Research assistants were well trained in conducting interviews, transcription, coding, as well in the technological execution of research. 
We stopped conducting studies after a clear picture of how the technical models and interest within the partner models developed was established. The rather small sample size was sufficient for our findings. Further, we also identified limitations regarding interests: Assumed interests in pre-interviews were ideas EXs had about what EEs would like, enjoy, or literally would be interested in. In video recall-interviews though, as explanations were compact and the explanandum straightforward, not all expressed interests \cite{ref_article45} were interests in a narrow sense but were occasionally directly linked to the aspect that was explained in a particular moment. Hopefully, with more complex artifacts, we can better differentiate between different forms of interests. Then, it could be interesting to assess more diverse aspects of motivation in the partner model. 
Regardless of the fact that in interviews with open-ended questions, interviewees could explore the topic and generate rich and meaningful answers, it needs to be addressed that answers that are given depended on the questions and behavior of interviewers. The introspective video-recall method brought numerous advantages in the form of detailed insights on the EXs’ developing assumptions regarding technical models and related interests of EEs in specific moments of the explanation. But the method also had disadvantages. Study participants could feel a certain degree of stress and anxiousness because of the research environment. Watched pre-selected video scenes also had limitations regarding video sections, video quality, and threshold of acoustic transmission \cite{ref_article5}. Therefore, we optimized technical solutions and ensured that the research environment was perceived as safe and welcoming.\\
\newline
\noindent \textbf{Future Work.} The findings of this qualitative study are meaningful and provide hints on implications and important further research in this direction. To increase comparability and reduce efforts, a quantitative research approach, which allows a bigger sample size, is recommended. A switch from the investigation of naturalistic explanations to an experimental research design, where aspects of the interaction with regard to Architecture and Relevance are varied, is intended. In a second next step the development of questionnaires is planned to assess the developing technical model from the EEs’ perspective.\\
Another logical research aim for the future would be a switch from analog technological artifacts to digital technological artifacts. The technological artifact we used was rather simple but was sufficient to learn more about the development of EEs’ technical models within partner models in terms of the dual nature of technological artifacts in naturalistic explanations. Research with this straightforward technological artifact already revealed that both sides of the duality were addressed and that both perspectives played a crucial role in the explanation. When conducting research with digital technological artifacts of higher complexity and depth, the technical model within the partner model might even have greater importance. The reasons could be found to varying degrees in prior knowledge, but particularly in differently accentuated interests. Imagine an artifact with diverse functions. Accordingly, EEs might only be interested in certain features (with their own structure and purpose) and might not need to develop in-depth knowledge regarding every single aspect. This was different in our investigations on Quarto, where almost all available information was required to be provided to enable EEs to understand and potentially apply their gained knowledge when playing the game. Thus, with artifacts of higher complexity, it is of interest how the development of technical models within partner models differs from those of rather simple artifacts. The EXs probably need to consider and monitor the behavior of the EEs and their needs even more closely. Again, the dual nature of technological artifacts might be useful as EEs can freely accentuate which features they are really interested in and from which perspective they demand an explanation. \\
\\
\noindent \textbf{Conclusion.} In this study, we empirically showed how the EXs’ assumptions about the EEs’ technical models and interests develop on multiple levels. It can be concluded that EXs were able to develop well-defined technical models and interests within the partner models, in which knowledge gaps and unsatisfied needs were also identified. The results are important for XAI research and development as the technical model and interests within the partner model played a critical role in the overall explanations. To not consider EEs and their needs in explanations should be avoided by humans as well as XAI. Hence, ideas for further development of XAI in the form of practical implications for synthesis of explanations were provided. Ideally, XAI will be enabled in the future to consider the knowledge and interests of users and adapt the explanations accordingly. The investigated technical model and interests in the partner model could serve as a base for the development of user models for person-specific and adaptive explainable systems. \\
\\
\small
\textbf{Acknowledgments.} This research was funded by the Deutsche Forschungsgemeinschaft (DFG, German Research Foundation): TRR 318/1 2021 - 438445824. \\
\\
\textbf{Disclosure of Interests.} This study was approved by the Paderborn University Ethics Board. All participants participated voluntarily and provided written informed consent.
\\
%
%
%
%

\end{document}